\documentclass{article}

\PassOptionsToPackage{numbers}{natbib}
\usepackage[final]{nips_2017}

\usepackage[utf8]{inputenc} 
\usepackage[T1]{fontenc}    
\usepackage{hyperref}       
\usepackage{url}            
\usepackage{booktabs}       
\usepackage{amsfonts}       
\usepackage{nicefrac}       
\usepackage{microtype}      
\usepackage{amssymb,amsmath,mathtools, mathrsfs, dsfont}
\usepackage{graphicx, epstopdf}
\usepackage{caption}

\def\kth{$k^{\text{th}}$}
\def\ith{$i^{\text{th}}$}	
\def\jth{$j^{\text{th}}$}

\renewcommand{\vec}[1]{\mathbf{#1}}

\newcommand{\KLD}[2]{D_{\text{KL}}\left( #1\parallel#2\right)}

\DeclareMathOperator*{\argmax}{arg\,max}

\title{Detection Algorithms for Communication Systems Using Deep Learning}

\author{
  Nariman Farsad\\
  Department of Electrical Engineering\\
  Stanford University\\
  Stanford, CA 94305 \\
  \texttt{nfarsad@stanford.edu} \\
  \And 
  Andrea Goldsmith \\
  Department of Electrical Engineering\\
  Stanford University\\
  Stanford, CA 94305 \\
  \texttt{andrea@wsl.stanford.edu} \\
}

\begin{document}

\maketitle

\begin{abstract}
The design and analysis of communication systems typically rely on the development of mathematical models that describe the underlying communication channel, which dictates the relationship between the transmitted and the received signals. However, in some systems, such as molecular communication systems where chemical signals are used for transfer of information, it is not possible to accurately model this relationship. In these scenarios, because of the lack of mathematical channel models, a completely new approach to design and analysis is required. In this work, we focus on one important aspect of communication systems, the detection algorithms, and demonstrate that by borrowing tools from deep learning, it is possible to train detectors that perform well, without any knowledge of the underlying channel models. We evaluate these algorithms using experimental data that is collected by a chemical communication platform, where the channel model is unknown and difficult to model analytically. We show that deep learning algorithms perform significantly better than a simple detector that was used in previous works, which also did not assume any knowledge of the channel.
\end{abstract}

\section{Introduction}
The design and analysis of communication systems typically rely on developing mathematical models that describe signal transmission, signal propagation, receiver noise, and many of the other components of the system that affect the end-to-end signal transmission. Particularly, most communication systems today lend themselves to tractable channel models based on a simplification of Maxwell's equations. However, there are cases where this does not hold, either because the electromagnetic (EM) signal propagation is very complicated and/or poorly understood, or because the signal is not an EM signal and hence good models for its propagation don't exist. 
Some examples of the latter includes underwater communication using acoustic signals \cite{sto09}, and a new technique called molecular communication, which relies on chemical signals to interconnect tiny devices with sub-millimeter dimensions in environments such as inside the human body \cite{mor06,aky08,eckBook,far16ST}. In many chemical communication systems, the underlying channel models are unknown and  it is difficult to derive tractable analytical models. For example, signal propagation in chemical communication channels with multiple reactive chemicals are highly complex \cite{reactDiffBook}. Therefore, an approach to design and engineer these systems that does not require analytical channel models is required. 

Machine learning and deep learning are techniques that could be used for this task \cite{lec15,goodfellowBook,ibn00}. Some examples of machine learning tools applied to design problems in communication systems include multiuser detection in code-division multiple-access (CDMA) systems \cite{aaz92,mit94,jua06,isi07}, decoding of linear codes \cite{nac16}, design of new modulation and demodulation schemes \cite{osh16}, and estimating channel model parameters \cite{lee17}. Most previous work have used machine learning and deep learning to improve one component of the communication system based on the knowledge of the underlying channel models. 

Our approach is different from these work since we assume that the mathematical models for the communication channel are completely unknown. Particularly, we focus on one of the important modules of any communication system, the detection algorithm, where the transmitted signal is estimated from a noisy and corrupted version that is observed at the receiver. We demonstrate that, using deep learning, it is possible to train a detector without any knowledge of the underlying channel models. In this scheme, the receiver goes through a training phase where a deep learning detector is trained using known transmission signals.  At first glance, this may seem like extra overhead. However,  any channel model needs to be validated experimentally before it can used for designing detection algorithms. In fact, even in wireless radio communication, experimental data are used to ``train'' and ``learn'' (i.e., validate and revise) the channel models. Another important benefit of using deep learning detectors is that they return likelihoods of each symbol, which is used in soft decision error correction decoders that are typically the next module in a communication system after the detector.

To evaluate the performance of deep learning detectors, we focus on molecular communication where chemical signals are used to transfer messages \cite{mor06,aky08,eckBook,far16ST}. Molecular communication is a perfect platform for evaluation since the underlying channel models are nonlinear and unknown. An experimental platform is used for collecting measurement data, which is used to train and evaluate the deep learning detection algorithms. We demonstrate that for this setup deep learning detectors perform significantly better than a simple detector that was used in previous works, which also did not assume any knowledge of the channel \cite{far13,koo16}. 

Note that although the proposed techniques can be used with any communication system, use of these techniques in chemical communication systems enable many interesting applications. For example, one particular area of interest is in-body communication where bio-sensors, such as synthetic biological devices, constantly monitor the body for different bio-markers for diseases \cite{ata12CM}. Naturally, these biological sensors, which are adapt at detecting biomarkers {\em in vivo} \cite{and06,dan15,slo15}, need to convey their measurements to the outside world. Chemical signaling is a natural solution to this communication problem where the sensor nodes chemically send their measurements to each other or to other devices under/on the skin. The device on the skin is connected to the Internet through wireless technology and can therefore perform complex computations. Thus, the experimental platform we use in this work to validate deep learning algorithms for signal detection can be used directly to support this important application. 



The rest of the paper is organized as follows. In Section \ref{sec:probState}, we present the problem statement. Then in Section \ref{sec:deepDetect}, detection algorithms based on deep learning are introduced. An experimental platform, which is used for data collection, is presented in Section \ref{sec:expSetup}. Section \ref{sec:results} compares the performance of different detection algorithms, and concluding remarks are provided in Section \ref{sec:conclusion}.

\section{Problem Statement}
\label{sec:probState}
In a digital communication system a message is converted into a sequence of transmission symbols that represent different transmission signals. Let $\mathcal{X}=\{s_1, s_2, \cdots, s_m\}$ be a finite set of all transmission symbols, and $x_k\in\mathcal{X}$ be the \kth~transmission symbol. The signal corresponding to the \kth~transmission propagates in the environment, and a noisy and corrupted version, which is represented by the random vector $\vec{y}_k$ of length $\ell$, is detected at the receiver. In a simple memoryless system, the relation between the transmitted symbol and the received signal can be represented by a statistical channel model that characterizes the probability distribution function $P(\vec{y}_k \mid x_k)$.

In many communication systems, the channel characteristics have memory, e.g., because of a person moving or in some system due to intersymbol interference (ISI) from previous transmissions. Therefore, the statistics of the channel also depend on the current state of the channel. In this work, we focus on channels with ISI. Note that chemical communication channels suffer from significant ISI. For the ISI channel, the relation between the transmitted signal and the received signal is characterized by the statistical channel model $P(\vec{y}_k, \vec{y}_{k-1}, \cdots, \vec{y}_1 \mid x_k, x_{k-1}, \cdots, x_1)$.

The receiver has no prior knowledge of the transmitted signal and a detection algorithm at the receiver is used to estimate the transmitted symbols from the received signals. Let $\hat{x}_k$ be the symbol estimated by the detection algorithm for the \kth~transmission. Then detection can be performed through symbol-by-symbol detection where $\hat{x}_k$ is estimated from $\vec{y}_k$, or using sequence detection where the sequence $\hat{x}_k, \hat{x}_{k-1}, \cdots, \hat{x}_1$ is estimated from the sequence $\vec{y}_k, \vec{y}_{k-1}, \cdots, \vec{y}_1$.  The traditional approach to designing the detection algorithm has relied on probabilistic models of the channel. For example, for a simple channel with no ISI, given by the channel model $P_{\text{model}}(\vec{y}_k \mid x_k)$, and a known probability mass function (PMF) for the transmission symbols $P_X(x)$ (note that we know this PMF since it is part of the system design), a maximum a posteriori estimation (MAP) algorithm can be devised as
\begin{align}
\label{eq:egMAP}
\hat{x}_k = \argmax_{x \in \mathcal{X}} P_{\text{model}}(\vec{y}_k \mid x)P_X(x).
\end{align}

One major challenge with this technique is that the channel model, and in particular its input-output statistics, has to be known. Moreover, the model has to be simple enough such that \eqref{eq:egMAP} can be solved in a computationally efficient manner \cite{mur17}. In many cases, this may be the case. For example, chemical communication systems may be nonlinear and the analytical channel models may be non-existent \cite{reactDiffBook,far16SPAWC}. Therefore, this begs the question as to how we can design detectors when the underlying channel models are unknown or are sufficiently complex such that devising computationally efficient detection algorithms may not be possible.

In the next section we show how deep learning can be used as a solution to this problem.    

\section{Detection Using Deep Learning}
\label{sec:deepDetect}
Estimating the transmitted symbol from the received signals $\vec{y}_k$ can be mapped to a clustering problem. Particularly, let $m=|\mathcal{X}|$ be the total number of symbols, and let $\vec{x}_k$ be the one-hot representation of the symbol transmitted during the \kth~transmission, given by
\begin{align}
\label{eq:actualPMF}
	\vec{x}_k = 
	\begin{bmatrix}
		\mathds{1}(x_k=s_1), 
		\mathds{1}(x_k=s_2), 
		\cdots, 
		\mathds{1}(x_k=s_m) 
	\end{bmatrix}^\intercal,
\end{align}
where $\mathds{1}(.)$ is the indicator function. Therefore, the element corresponding to the symbol that is transmitted is 1, and all other elements of $\vec{x}_k$ are zero. Note that this is also the PMF of the transmitted symbol during the \kth~transmission where at the transmitter, with probability 1, one of the symbols is transmitted.

Designing the detection algorithm involves two phases. In the first phase, known sequences of transmission symbols are transmitted repeatedly and received by the system to create a set of training data. For example, random number generators with the same seed at the transmitter and receiver could be used for this task. Let $\vec{X}_K = [\vec{x}_1,\vec{x}_2,\cdots,\vec{x}_K]$ be a sequence of $K$ consecutively transmitted symbols, and $\vec{Y}_K = [\vec{y}_1,\vec{y}_2,\cdots,\vec{y}_K]$ the corresponding sequence of received feature vectors. Then, the training dataset is represented by
\begin{align}
	\{(\vec{X}^{(1)}_K,\vec{Y}^{(1)}_K),(\vec{X}^{(2)}_K,\vec{Y}^{(2)}_K), \cdots, (\vec{X}^{(n)}_K,\vec{Y}^{(n)}_K) \},
\end{align}
which consists of $n$ samples of $K$ consecutive transmissions. 

This dataset is then used to train a deep learning classifier that classifies the received signal $\vec{y}_k$ as one of the transmission symbols in $\mathcal{X}$. The input to the deep learning network can be the received signals $\vec{y}_k$, or a set of features $\vec{r}_k$ extracted from the received signals. The deep learning network output are the vectors $\hat{\vec{x}}_k$, which is the estimated PMF that the \kth~transmission symbol belongs to each of the $m$ possible symbols. Note that this output is also useful for soft decision error correction decoders, which are typically the next module after detection. The symbol with the highest mass point in $\hat{\vec{x}}_k$ is chosen as the estimated symbol for the \kth~transmission. 

During the training, an optimization algorithm such as stochastic gradient decent is used to minimize the loss between the actual PMF $\vec{x}_k$ in \eqref{eq:actualPMF}, and the estimated PMF $\hat{\vec{x}}_k$ \cite{goodfellowBook}. Particularly, in this work we use the categorical cross-entropy loss function $\mathcal{L}$ for this optimization \cite{goodfellowBook}. For symbol-by-symbol detection, this function is defined as
\begin{align}
\label{eq:lossSymbSymb}
\mathcal{L_{\text{symb}}} = H(\vec{x}_k,\hat{\vec{x}}_k) = H(\vec{x}_k) + \KLD{\vec{x}_k}{\hat{\vec{x}}_k},
\end{align} 
and for sequence detection, where a sequence of length $\tau$ is estimated from the received signals, it is given by
\begin{align}
\label{eq:lossSeqSeq}
\mathcal{L_{\text{seq}}} = \sum_{k=1}^{\tau}H(\vec{x}_k,\hat{\vec{x}}_k) = \sum_{k=1}^{\tau}H(\vec{x}_k) + \KLD{\vec{x}_k}{\hat{\vec{x}}_k},
\end{align} 
where $H(\vec{x}_k,\hat{\vec{x}}_k)$ is the cross entropy between the correct PMF and the estimated PMF, and $\KLD{.}{.}$ is the Kullback-Leibler divergence \cite{cover-book}.  Note that minimizing the loss is equivalent to minimizing the cross-entropy or the Kullback-Leibler divergence distance between the true PMF given in \eqref{eq:actualPMF} and the one estimated based on the neural network. Alternatively, \eqref{eq:lossSymbSymb} and \eqref{eq:lossSeqSeq} can be written as
\begin{align}
\mathcal{L_{\text{symb}}} &= -\log\bigg(\hat{\vec{x}}_k [x_k]\bigg),\label{eq:likelihoodSymb}\\
\mathcal{L_{\text{seq}}} &= -\sum_{k=1}^{\tau} \log\bigg(  \hat{\vec{x}}_k [x_k]\bigg), \label{eq:likelihoodSeq}
\end{align} 
where $\hat{\vec{x}}_k [x_k]$ indicates the element of $\hat{\vec{x}}_k$ corresponding to symbol that was actually transmitted, i.e., $x_k$. Note that minimizing \eqref{eq:likelihoodSymb} and \eqref{eq:likelihoodSeq} is equivalent to maximizing the log expression, which essentially maximizes the log-likelihoods. Therefore, during the training, known transmission data are used to train a detector that {\em maximizes log-likelihoods}.  Using Bayes' theorem, it is easy to show that minimizing the loss in \eqref{eq:likelihoodSymb} is equivalent to maximizing \eqref{eq:egMAP}. Therefore, deep learning can be a powerful tool for designing detection algorithms for communication systems, especially when the underlying channel models are unknown. We now discuss several generic architectures that can be used for symbol-by-symbol and sequence detection.

\begin{figure}[t]
	\begin{center}
		\includegraphics[width=1\textwidth,keepaspectratio]{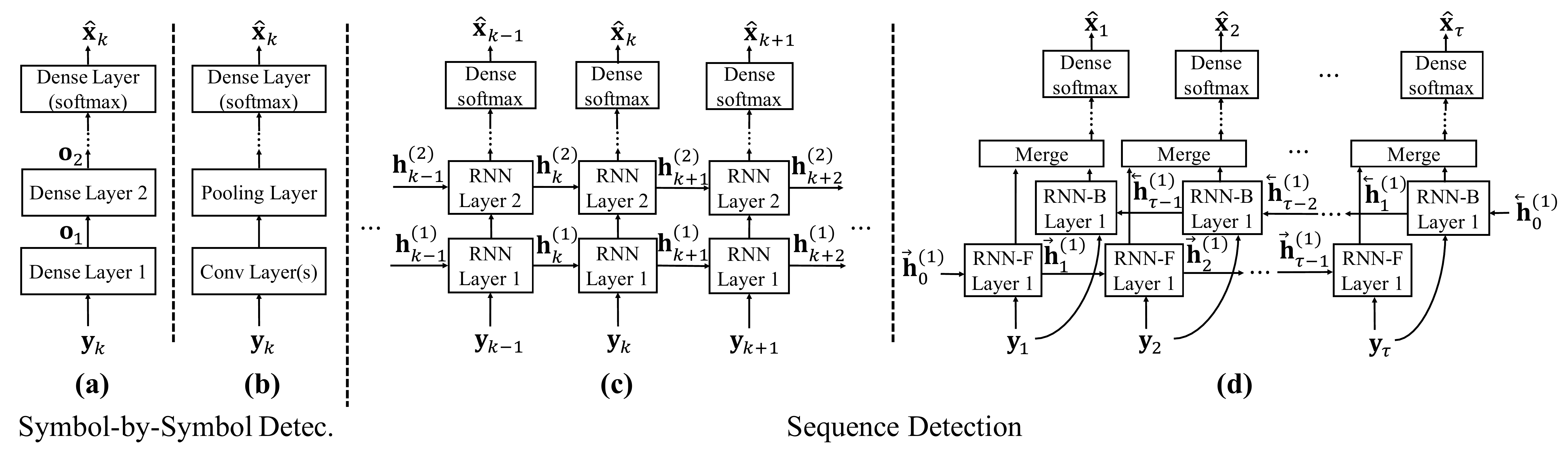}
	\end{center}
	\caption{\label{fig:NNarchitecture} Different neural network architectures for detection.}
\end{figure}
\subsection{Symbol-by-Symbol Detection}
The most basic neural network architecture that can be employed for detection uses several dense neural network layers followed by a final softmax layer \cite{lec15,goodfellowBook}. In this network, the \ith~layer is given by
\begin{align}
	\label{eq:denseLayer}
	\vec{o}_i = f(\vec{W}_i \vec{o}_{i-1} + \vec{b}_i),
\end{align}
where $f()$ is the activation function, $\vec{W}_i$ and $\vec{b}_i$ are the weight and bias parameters, and $\vec{o}_i$ is the output of the \ith~layer. The input to the first layer is the received signal $\vec{y}_k$ or the feature vector $\vec{r}_k$, which is selectively extracted from the received signal through preprocessing. The output of the final layer is of length $m$ (i.e., the cardinality the symbol set), and the activation function for the final layer is the softmax activation given by
\begin{align}
	\label{eq:softmax}
	\phi_i(\vec{z})=\frac{e^{z_i}}{\sum_j e^{z_j}}.
\end{align} 
This ensures that the output of the layer $\hat{\vec{x}}_k$ is a PMF. Figure \ref{fig:NNarchitecture}(a) shows the structure of this neural network. 

In many cases, the received signal may be randomly shifted in time because of random delays introduced by the communication channel. A more sophisticated neural network that is more resilient to this effect is the convolution neural network (CNN) \cite{law97,kri12,lec15}. Essentially, for our purposes the CNN is a set of filters that extracts the most relevant features from the data. The final layer in CNN detector is a dense layer with output of length $m$, and a softmax activation function, which result in an estimated $\hat{\vec{x}}_k$. Figure \ref{fig:NNarchitecture}(b) shows the structure of this neural network.   

For symbol-by-symbol detection the estimated PMF $\hat{\vec{x}}_k$ is given by
\begin{align}
\label{eq:estSymbPMF}
\hat{\vec{x}}_k = 
\begin{bmatrix}
\hat{P}_{\text{model}}(\hat{x}_k=s_1|\vec{y}_k),
\hat{P}_{\text{model}}(\hat{x}_k=s_2|\vec{y}_k), 
\cdots,
\hat{P}_{\text{model}}(\hat{x}_k=s_m|\vec{y}_k)
\end{bmatrix}^\intercal,
\end{align}
where $\hat{P}_{\text{model}}$ is the probability of estimating each symbol based on the neural network model used. The better the structure of the neural network at capturing the physical channel characteristics, the better this estimate and the results. Therefore, it is important to use insights from the physical channel characteristics when designing the network architecture.   

\subsection{Sequence Detection}
The symbol-by-symbol detector may not perform well in systems with ISI and memory. In this case, sequence detection can be performed using recurrent neural networks (RNN) \cite{lec15,goodfellowBook}. In particular, we use long short-term memory (LSTM) networks in this work \cite{hoc97}, which is composed of the following
\begin{align}
	\vec{i}_k &= \sigma(\vec{W}_{y,i}\vec{y}_k+\vec{W}_{a,i}\vec{a}_{k-1}+\vec{W}_{c,i}\vec{c}_{k-1}+\vec{b}_i),\label{eq:LSTMnode1}\\
	\vec{f}_k &= \sigma(\vec{W}_{y,f}\vec{y}_k+\vec{W}_{a,f}\vec{a}_{k-1}+\vec{W}_{c,f}\vec{c}_{k-1}+\vec{b}_f),\label{eq:LSTMnode2}\\
	\vec{c}_k &= \vec{f}_k \odot \vec{c}_{k-1} + \vec{i}_k  \odot \tanh (\vec{W}_{y,c}\vec{y}_k+\vec{W}_{a,c}\vec{a}_{k-1}+\vec{b}_c), \label{eq:LSTMnode3}\\
	\vec{u}_k &= \sigma(\vec{W}_{y,u}\vec{y}_k+\vec{W}_{a,u}\vec{a}_{k-1}+\vec{W}_{c,u}\vec{c}_{k}+\vec{b}_u), \label{eq:LSTMnode4}\\
	\vec{a}_k &= \vec{u}_k \odot \tanh(\vec{c}_k), \label{eq:LSTMnode5}
\end{align}
where $ \odot$ is the element-wise vector multiplication, $\vec{W}$ and $\vec{b}$ are weights and biases, and $\sigma$ is the logistic sigmoid function \cite{goodfellowBook}. The output of the LSTM cell is $\vec{a}_k$, and the hidden state $\vec{h}_k$ (see Figure \ref{fig:NNarchitecture}(c)), which is passed to the next cell in the next time instance, are $\vec{c}_k$ and $\vec{a}_k$. The output of the LSTM may be passed to other LSTM layers as shown in Figure \ref{fig:NNarchitecture}(c). Let $\vec{a}^{(l)}_k$ be the output of the final LSTM layer. Then a dense layer with the softmax activation in \eqref{eq:softmax} is used to obtain $\hat{\vec{x}}_k$ as
\begin{align}
	\hat{\vec{x}}_k = \phi(\vec{W}_{a} \vec{a}^{(l)}_k + \vec{b}_x).
\end{align}
This estimated $\hat{\vec{x}}_k$ is essentially a PMF given by
\begin{align}
\label{eq:estForwSeqPMF}
\hat{\vec{x}}_k = 
\begin{bmatrix}
\hat{P}_{\text{model}}(\hat{x}_k=s_1|\vec{y}_k,\vec{a}^{(l)}_{k-1},\vec{c}^{(l)}_{k-1}) \approx \hat{P}_{\text{model}}(\hat{x}_k=s_1|\vec{y}_k,\vec{y}_{k-1},\cdots,\vec{y}_1)   \\
\hat{P}_{\text{model}}(\hat{x}_k=s_2|\vec{y}_k,\vec{a}^{(l)}_{k-1},\vec{c}^{(l)}_{k-1}) \approx \hat{P}_{\text{model}}(\hat{x}_k=s_2|\vec{y}_k,\vec{y}_{k-1},\cdots,\vec{y}_1) \\
\vdots \\
\hat{P}_{\text{model}}(\hat{x}_k=s_m|\vec{y}_k,\vec{a}^{(l)}_{k-1},\vec{c}^{(l)}_{k-1}) \approx \hat{P}_{\text{model}}(\hat{x}_k=s_m|\vec{y}_k,\vec{y}_{k-1},\cdots,\vec{y}_1) 
\end{bmatrix},
\end{align}    
where $\hat{P}_{\text{model}}$ is the probability of estimating each symbol based on the neural network model used. Note that the decoder considers the information from previously received signals, encoded in $\vec{a}^{(l)}_{k-1}$ and $\vec{c}^{(l)}_{k-1}$,  as well as the received signal from current symbol for detection.

One issue with this detector is that the signal that is received in the \jth~transmission $\vec{y}_j$ where $j>k$ may carry information about the \kth symbol $x_k$, and this signal is not considered by the detector. One way to overcome this limitation is by using bidirectional RNNs (BRNNs) \cite{sch97} on sequences of fixed length $\tau$ (see Figure \ref{fig:NNarchitecture}(d)). Particularly, we use a bidirectional LSTM network \cite{gra05}. In this network, the sequence of received signals are once fed in the forward direction into one LSTM cell \eqref{eq:LSTMnode1}-\eqref{eq:LSTMnode5} resulting in output $\overrightarrow{\vec{a}}_k$, and once fed in backwards into another LSTM cell resulting in the output $\overleftarrow{\vec{a}}_k$. The two outputs may be passed to more bidirectional layers. Let $\overrightarrow{\vec{a}}^{(l)}_k$ and $\overleftarrow{\vec{a}}^{(l)}_k$ be the outputs of the final bidirectional layer. Then a dense layer with softmax activation is used to obtain $\hat{\vec{x}}_k$ as
\begin{align}
\hat{\vec{x}}_k = \phi(\vec{W}_{\overrightarrow{\vec{a}}} \overrightarrow{\vec{a}}^{(l)}_k + \vec{W}_{\overleftarrow{\vec{a}}} \overleftarrow{\vec{a}}^{(l)}_k+\vec{b}_x).
\end{align}
This estimated $\hat{\vec{x}}_k$ is essentially a PMF given by
\begin{align}
\label{eq:estBiDirSeqPMF}
\hat{\vec{x}}_k = 
\begin{bmatrix}
\hat{P}_{\text{model}}(\hat{x}_k=s_1|\vec{y}_k,\mspace{-3mu}\overrightarrow{\vec{a}}^{(l)}_{k-1},\mspace{-3mu}\overrightarrow{\vec{c}}^{(l)}_{k-1},\mspace{-3mu}\overleftarrow{\vec{a}}^{(l)}_{k-1},\mspace{-3mu}\overleftarrow{\vec{c}}^{(l)}_{k-1})\mspace{-3mu} \approx\mspace{-3mu} \hat{P}_{\text{model}}(\hat{x}_k=s_1|\vec{y}_\tau,\mspace{-3mu}\vec{y}_{\tau-1},\mspace{-3mu}\cdots,\mspace{-3mu}\vec{y}_1) \mspace{-3mu} \\
\hat{P}_{\text{model}}(\hat{x}_k=s_2|\vec{y}_k,\mspace{-3mu}\overrightarrow{\vec{a}}^{(l)}_{k-1},\mspace{-3mu}\overrightarrow{\vec{c}}^{(l)}_{k-1},\mspace{-3mu}\overleftarrow{\vec{a}}^{(l)}_{k-1},\mspace{-3mu}\overleftarrow{\vec{c}}^{(l)}_{k-1})\mspace{-3mu} \approx\mspace{-3mu} \hat{P}_{\text{model}}(\hat{x}_k=s_2|\vec{y}_\tau,\mspace{-3mu}\vec{y}_{\tau-1},\mspace{-3mu}\cdots,\mspace{-3mu}\vec{y}_1) \mspace{-3mu}\\
\vdots \\
\hat{P}_{\text{model}}(\hat{x}_k=s_m|\vec{y}_k,\mspace{-3mu}\overrightarrow{\vec{a}}^{(l)}_{k-1},\mspace{-3mu}\overrightarrow{\vec{c}}^{(l)}_{k-1},\mspace{-3mu}\overleftarrow{\vec{a}}^{(l)}_{k-1},\mspace{-3mu}\overleftarrow{\vec{c}}^{(l)}_{k-1})\mspace{-3mu} \approx\mspace{-3mu} \hat{P}_{\text{model}}(\hat{x}_k=s_m|\vec{y}_\tau,\mspace{-3mu}\vec{y}_{\tau-1},\mspace{-3mu}\cdots,\mspace{-3mu}\vec{y}_1) \mspace{-3mu}
\end{bmatrix}.
\end{align}
Therefore, the current symbol is decoded based on the signal received during the current symbol, as well as all the other signals received over the sequence of length $\tau$, which are encoded in $\overrightarrow{\vec{a}}^{(l)}_{k-1}, \overrightarrow{\vec{c}}^{(l)}_{k-1}, \overleftarrow{\vec{a}}^{(l)}_{k-1}$, and $\overleftarrow{\vec{c}}^{(l)}_{k-1}$. 

To evaluate the performance of the deep learning detection algorithms, we use an experimental platform to collect data for training and testing the algorithms. In the next section we describe this platform.


%
%
%

\vspace{-0.25cm}
\section{Experimental Setup}
\label{sec:expSetup}
\vspace{-0.25cm}
We consider a chemical communication system for evaluating the performance of deep learning detection algorithms. There are several benefits to using this system. First, the experimental platform is inexpensive and can be used to collect large amounts of data required for training and testing. Second, chemical communication channels are poorly understood and the underlying end-to-end channel models are unknown. Third, the multi-chemical signal propagation, where multiple reactive chemical are present in the environment, tend to be inherently nonlinear, which makes deep learning detection an attractive solution. Finally, for the envisioned application of in-body chemical communication, where biological sensing devices \cite{and06,dan15,slo15} monitor the body and chemically send their measurements to a device on or under the skin, each person may have different body chemistry, which may require detection algorithms trained on each individual. 

\begin{figure}
	\begin{center}
		\includegraphics[width=1\textwidth,keepaspectratio]{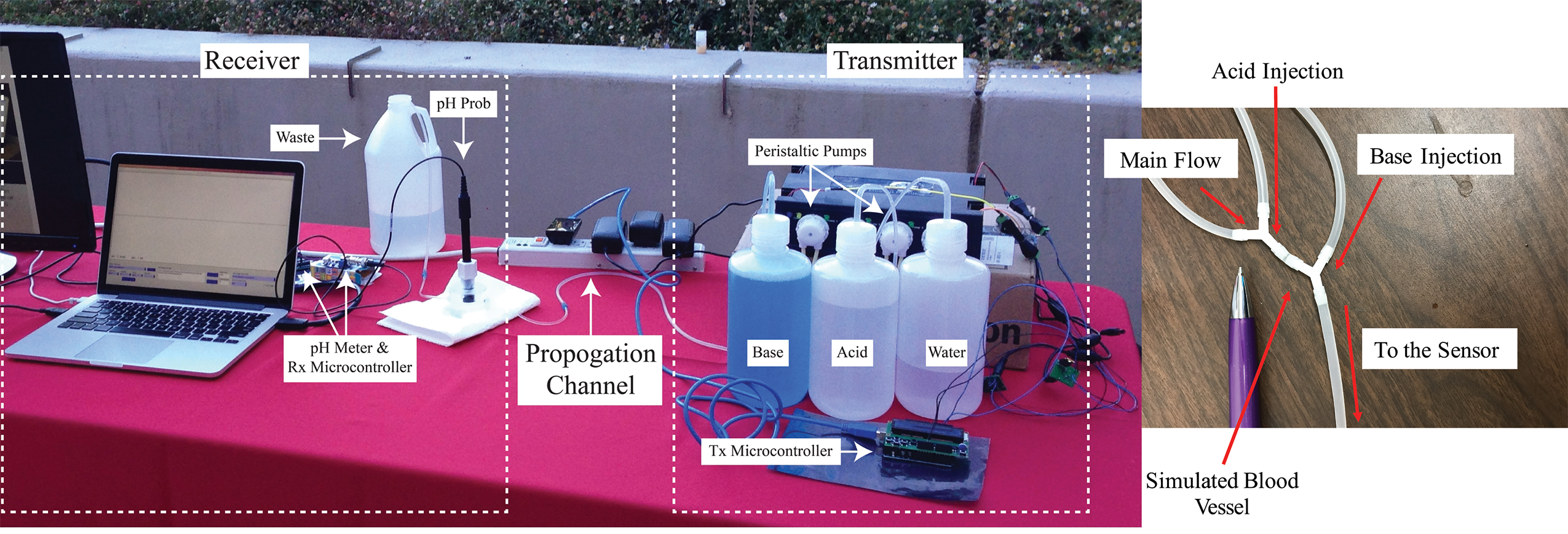}
	\end{center}
	\vspace{-0.3cm}
	\caption{\label{fig:thePlatform} The experimental platform.}
	\vspace{-0.5cm}
\end{figure}
Figure \ref{fig:thePlatform} shows a demonstrator for a single-link point-to-point in-vessel communication system. The platform uses peristaltic pumps to inject different chemicals into a main fluid flow in small silicon tubes. Multiple tubes with different diameters can be networked in branches to replicate a more complex environment such as the cardiovascular system in the body or complex networks of pipes in industrial complexes and city infrastructures. In the platform, there is always a main fluid flow in the tubes, for example, water or blood. The transmitter uses peristaltic pumps to inject different reactive chemicals into the main flow. Some examples includes acids and bases, or proteins, carbohydrates and enzymes. The central receiver uses a sensor such as a pH electrode or a glucose sensor to detect the chemical signals transmitted by the sender. In our platform, the main fluid flow is water and the transmitter used acids (vinegar) and bases (window cleaning solution) to encode information on the pH level. We use these particular chemicals since they are easily available and inexpensive. However, the results can be extended to blood as the main flow, and proteins and enzymes as the chemicals that are released by the transmitter.

In the platform, time-slotted communication is employed where the transmitter modulates information on acid and base signals by injecting these chemicals into the channel during each symbol duration. We use a simple binary modulation in this work where the bit-0 is transmitted by pumping acid into the environment for 30 ms at the beginning of the symbol interval, and bit-1 is represented by pumping base into the environment for 30 ms at the beginning of the symbol interval. The symbol interval then depends on the period of silence that follows the 30 ms chemical injection. In particular, four different pause durations of 220 ms, 304 ms, 350 ms, and 470 ms are used in this work to represent bit rates of 4, 3, 2.6, and 2 bps.

\begin{figure*}[t]
	\normalsize
	\centering
	\begin{minipage}{.43\textwidth}
		\begin{center}
			\includegraphics[width=\columnwidth,keepaspectratio]{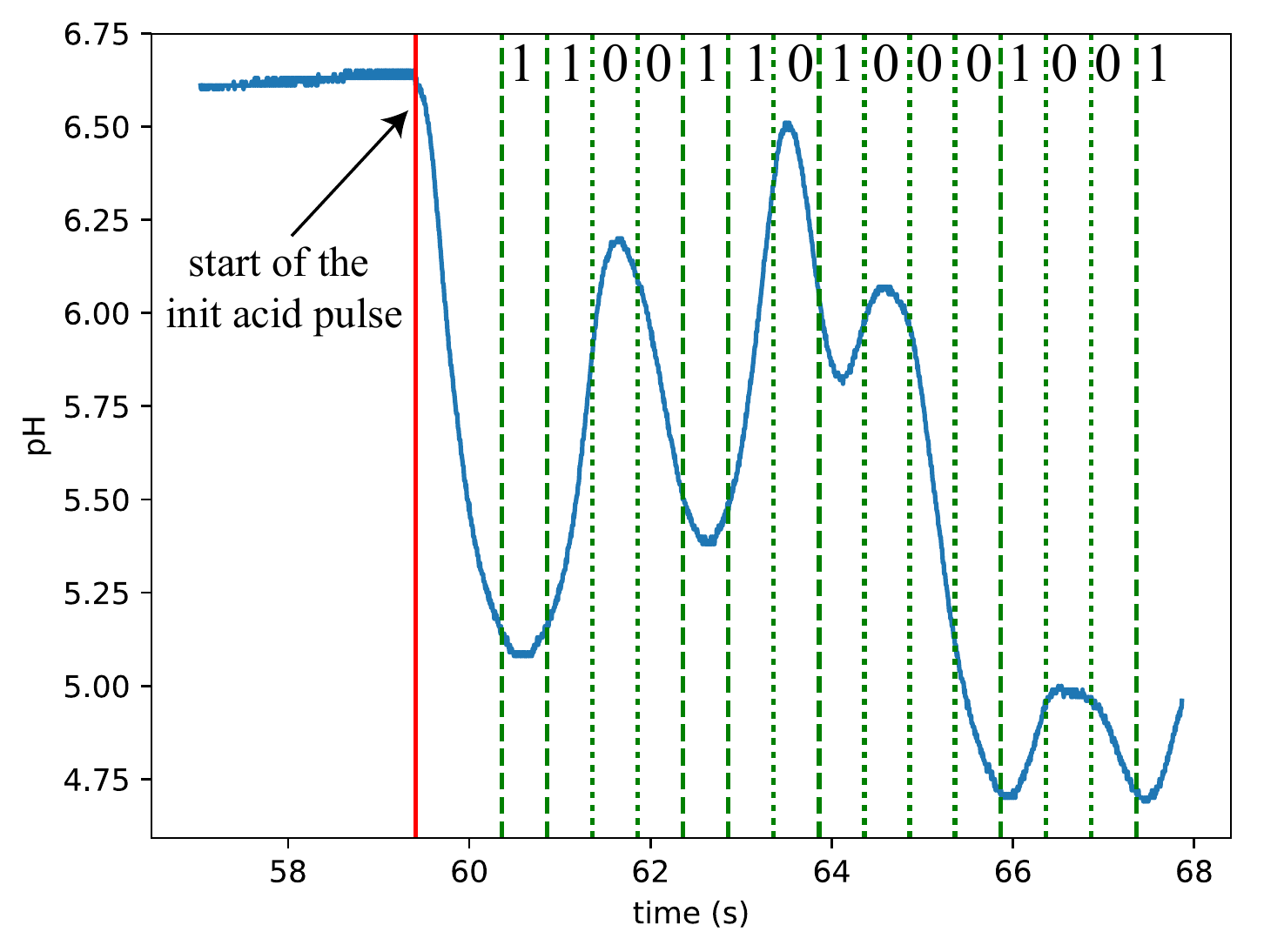}
		\end{center}
\caption{\label{fig:sampleTrans} The received pH signal for a bit sequence transmission.}
	\end{minipage}
	\hspace{0.8cm}
	\begin{minipage}{.43\textwidth}
		\vspace{-0.4cm}
		\begin{center}
			\includegraphics[width=\columnwidth,keepaspectratio]{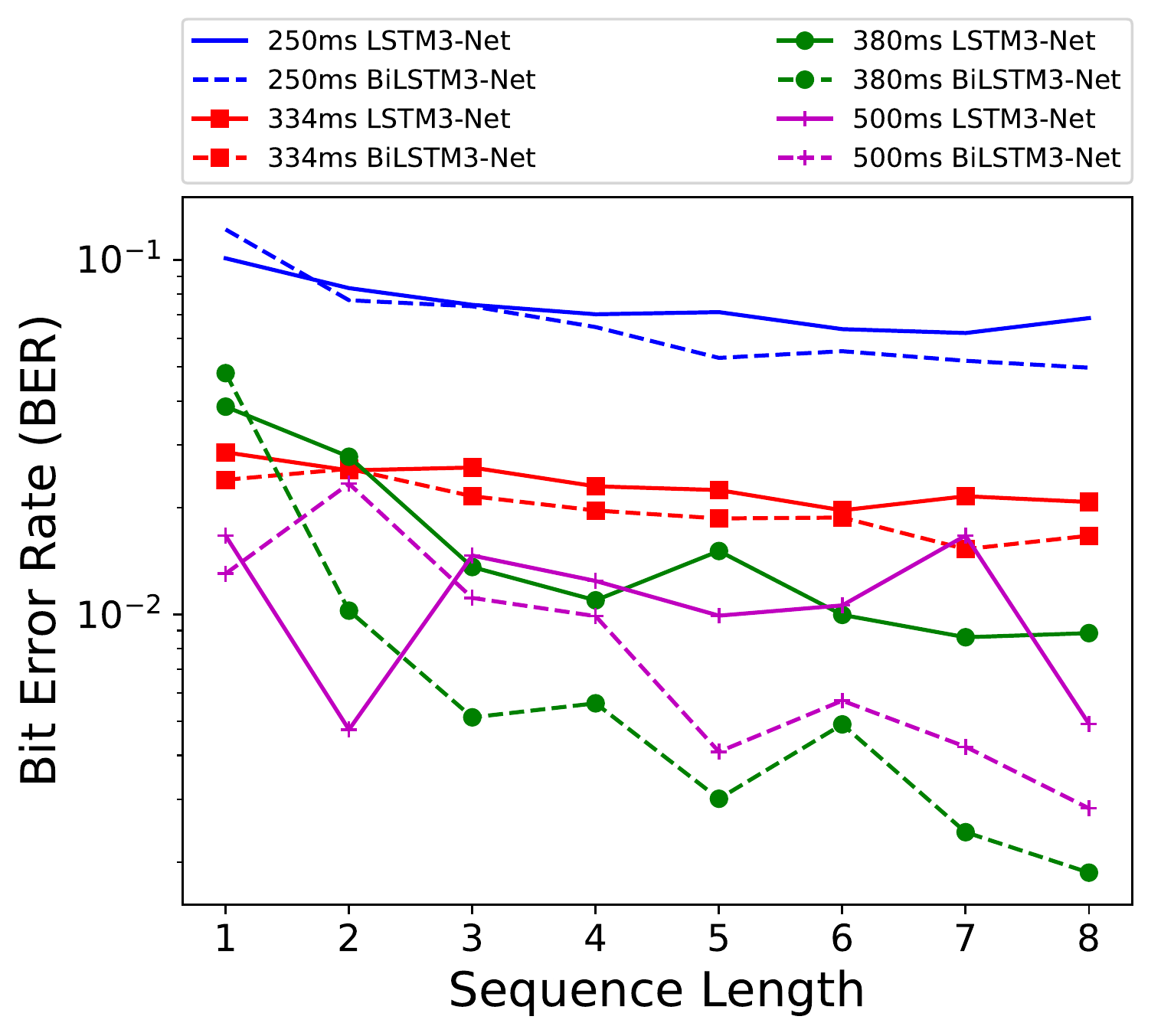}
		\end{center}
		\vspace{-0.4cm}
		\caption{\label{fig:BERvsSeqLen} Bit error rate versus the sequence length in sequence detectors.}
	\end{minipage}
\vspace{-0.6cm}
\end{figure*}

To synchronize the transmitter and the receiver, every message sequence starts with one initial injection of acid into the environment for 100 ms, followed by 900 ms of silence. The receiver then detects the starting point of this pulse and uses it to synchronize itself with the transmitter. Figure~\ref{fig:sampleTrans} shows the received pH signal for the transmission sequence ``110011010001001''. The start of the initial acid pulse detected by the receiver is shown using the red line. This detected time is used for synchronization and all the subsequent symbol intervals are shown by the green dashed and dotted lines. The dashed lines are used to indicate a bit-1 transmission and dotted lines to indicate a bit-0. A special termination sequence can be used to indicate the end of transmission. For example, if 5-bit encoded letters are used \cite{far13}, the all zero sequence can be used to signal the end of transmission.

Although it is difficult to obtain analytical models for multi-chemical communication systems \cite{reactDiffBook}, it is expected that when an acid pulse is transmitted, the pH should drop, and when a base pulse is injected into the environment, the pH should increase. Therefore, one approach to detection is to use the rate of change of pH to detect the symbols. To remove the noise from the raw pH signal, we divide the symbol interval (the time between green lines in  Figure~\ref{fig:sampleTrans}) into a number of equal subintervals or bins. Then the pH values inside each bin are averaged to represent the pH value for the corresponding bin. Let $B$ be the number of bins, and $\vec{b}=[b_1,b_2,\cdots,b_B]$ the corresponding values of each bin. The difference between values of these bins $\vec{d}=[d_1,d_2,\cdots,d_{B-1}]$, where $d_{i-1} = b_i- b_{i-1}$, are used as a baseline detection algorithm. This algorithm has two parameters: the number of bins $B$, and the index $\gamma$ that is used for detection. If $d_\gamma\leq0$, acid transmission  and hence bit-0 is detected, and if $d_\gamma>0$, bit-1 is detected. This technique was used in previous work for detection \cite{far13,koo16}. In the next section we compare this to the deep learning detection algorithms presented in previous sections.

\vspace{-0.25cm}
\section{Results}
\label{sec:results}
\vspace{-0.25cm}
We use our experimental platform to collect measurement data and create the data set that is used for training and testing the detection algorithms. Particularly, as explained in the previous section, four symbol intervals of 250 ms, 334 ms, 380 ms and 500 ms are considered which results in data rates ranging from 2 to 4 bits per second (bps). For each symbol interval, random bit sequences of length 120 are transmitted 100 times, where each of the 100 transmissions are separated in times. This results in 12k bits per symbol interval that is used for training and testing. From the data, 84 transmissions per symbol interval (10,080 bits) are used for training and 16 transmissions are used for testing (1,920 bits). Therefore, the total number of training bits are 40,320, and the total number of bits used for testing is 7,680.  

We start by considering the baseline detection using the rate of change of the pH. We use the training data to find the best detection parameters $B$ and $\gamma$, and the test data for evaluating the performance. Besides this algorithm we consider different deep learning detectors. For all training, the Adam optimization algorithm \cite{kin14} is used with the learning rate $10^{-3}$.  Unless specified otherwise the number of epoch used during training is 200 and the batch size is 10.

We use two symbol-by-symbol detectors based on deep learning. The first detector uses three dense layers with 80 hidden nodes and a final softmax layer for detection. Each dense layer uses the rectified linear unit (ReLU) activation function \cite{goodfellowBook}. The input to the network are a set of features extracted from the received signal, which are chosen based on performance and the characteristics of the physical channel. The input includes: $b_1$ and $b_B$, i.e., the pH level in the first and the last bins, $\vec{d}$, i.e., the vector of differences of consecutive bins, and a number that indicates the symbol duration. Here, we refer to this network as {\em Dense-Net}. A second symbol-by-symbol detector uses 1-dimensional CNNs. Particularly, the best network architecture that we found has the following layers. 1) 16 filters of length 2 with ReLU activation; 2) 16 filters of length 4 with ReLU activation; 3) max pooling layer with pool size 2; 4) 16 filters of length 6 with ReLU activation; 5) 16 filters of length 8 with ReLU activation; 6) max pooling layer with pool size 2; 7) flatten and a softmax layer. The stride size for the filters is 1 in all layers. The input to this network is the vector of pH values corresponding to each bin $\vec{b}$. We refer to this network as {\em CNN-Net}.

For the sequence detection, we use three networks, two based on the architecture in Figure \ref{fig:NNarchitecture}(c) and one based on \ref{fig:NNarchitecture}(d). The first network has 3 LSTM layers and a final softmax layer, where the length of the output of each LSTM layer is 40. Two different inputs are used with this network. In the first, the input is the same set of features as the Dense-Net above. We refer to this network as {\em LSTM3-Net}. In the second, the input is the pretrained CNN-Net described above without the top softmax layer. In this network, the CNN-Net chooses the features directly from the pH levels of the bins. We refer to this network as {\em CNN-LSTM3-Net}. Finally, we consider three layers of bidirectional LSTM cells, where each cell's output length is 40, and a final softmax layer. The input to this network are the same set of features used for Dense-Net and the LSTM3-Net. We refer to this network as {\em BiLSTM3-Net}. 

For each of the networks, one parameter of interest that affects the input to the networks is the number of bins $B$. We have trained each network using different bin numbers to find the best value for each network. For the Dense-Net $B=9$, for the CNN-Net $B=30$ and for all networks where the first layer is an LSTM cell $B=8$. Note that during the training, for all deep learning detectors, the data from all symbol durations are used to train a single network, which can then perform detection on all symbol durations. 

We first compare the bit error rate (BER) performance of LSTM3-Net with BiLSTM3-Net as a function of sequence length. Figure \ref{fig:BERvsSeqLen} shows this result. Note that the number of epoch used for this plot is 50 and batch size 32. Since BiLSTM3-Net considers the received signal over the whole sequence for detection, on average, it performs better than LSTM3-Net. However, one issue with BiLSTM3-Net is that it must wait for the whole sequence to arrive before detection, whereas LSTM3-Net can detect each symbol as it arrives. Generally, for both networks the bit error rate drops as the sequence length increases.

\begin{table}[t]
	\scriptsize 
	\vspace{-0.6cm}
	\caption{Bit Error Rate Performance}
	\label{tb:BER}
	\centering
	\begin{tabular}{cccccccc}
		\toprule
		Symb. Dur. & Baseline  & Dense-Net & CNN-Net & LSTM3-Net8 & BiLSTM3-Net8 & LSTM3-Net120 & CNN-LSTM3-Net120  \\
		\midrule
		250 		& 0.1297	& 0.1057  & 0.1068 & 0.0685 & 0.0496 & {\bf 0.0333} & 0.0677  \\
		334     	& 0.0755	& 0.0245  & 0.0750 & 0.0207 & {\bf 0.0167} & 0.0417 & 0.0271  \\
		380     	& 0.0797    & 0.0380  & 0.0589 & 0.0088 & {\bf 0.0019} & 0.0083 & 0.0026  \\
		500     	& 0.0516    & 0.0115  & 0.0063 & 0.0049 & 0.0028 & {\bf 0.0005} & 0.0021  \\
		\bottomrule
	\end{tabular}
	\vspace{-0.7cm}
\end{table}
Table \ref{tb:BER} BER performance we were able to obtain for all detection algorithms, including the baseline algorithm. The number in front of the sequence detectors, indicates the sequence length. For example, LSTM3-Net120 is an LSTM3-Net that is trained on 120 bit sequences. In general, algorithms that use sequence detection perform significantly better than any symbol-by-symbol detection algorithm including the baseline algorithm. This is due to significant ISI present in chemical communication systems. 
Without knowing the channel models, it is difficult to devise sophisticated detection algorithms. This demonstrates the effectiveness of deep learning detectors in communication systems.    

\vspace{-0.25cm}
\section{Conclusions}
\label{sec:conclusion}
\vspace{-0.25cm} 
We used several deep learning architectures for building detectors for communication systems. Different architectures were considered for symbol-by-symbol detection as well as sequence detection. These algorithms have many potentials in systems where the underlying physical models of the channel are unknown or inaccurate. For example, molecular communication, which has many potential applications in medicine is very difficult to model using tractable analytical models. We use an experimental platform that simulates in-vessel chemical communication to collect experimental data for training and testing deep learning algorithms. We show that deep learning sequence detectors can improve the detection performance significantly compared to a baseline approach that also does not rely on channel models. This demonstrates the promising performance deep learning detection algorithms could have in designing future communication systems. Some interesting open problems that we would like to explore in the future are as follows. First, we would like to perform more parameter tuning to find better neural networks for detection. Another important problem we plan to explore is how resilient deep learning detectors are to changing channel conditions, for example, as the concentration of acid and base is changed, and what would be good protocols for quickly retraining the network when a change in channel conditions is detected. Finally, we plan to collect more data on our platform, and make our dataset publicly available to other researchers.

%



\small
\bibliographystyle{IEEEtran}
\bibliography{IEEEabrv,MolCom-Nariman,ML-Nariman}

\end{document}